\documentclass[10pt, a4paper]{article}

\usepackage[]{acl} 

\usepackage{times}
\usepackage{latexsym}

\usepackage[T1]{fontenc}

\usepackage[utf8]{inputenc}

\usepackage{microtype}

\usepackage{inconsolata}

\usepackage{graphicx}

\title{ESsEN: Training Compact Discriminative Vision-Language \\ Transformers in a Low-Resource Setting}

\author{Clayton Fields \\
Department of Computer Science \\
Boise State University\\
  \texttt{claytonfields@u.}\\
  \texttt{boisestate.edu}\\
      \And
 Casey Kennington \\
 Department of Computer Science \\
Boise State University\\
  \texttt{caseykennington}\\
  \texttt{boisestate.edu}\\ }

\begin{document}

\maketitle

\begin{abstract}
Vision-language modeling is rapidly increasing in popularity with an ever expanding list of available models. In most cases, these vision-language models have parameters in the tens of billions, which is necessary for some needs, but in many cases smaller models are necessary (e.g., on edge devices or independent robotic platforms). Unfortunately, there is little research in producing light-weight models or in training them with small datasets. Inspired by the language learning progression and data sparsity in child development, in this paper, we address both of these goals in a systematic fashion. We show that two-tower encoder models are superior to one-tower encoders in low-resource settings for discriminative English tasks. We show also that incorporating traditional convolutional networks into the two-tower transformer architecture can help produce parameter efficient vision-language models. Finally, we show that the cross-modal fusion module of two-tower encoders can vary significantly in shape and size while producing the same results. In addition, we present ESsEN, a compact vision-language model that can be trained end-to-end with relatively few resources that performs as well on several tasks with only a fraction of the parameters compared to other models. The experimental results and the tools we present here make vision-language modeling more accessible to a wider variety of researchers.
\end{abstract}

\section{Introduction}
\label{sec:introduction}

The adaptation of the transformer model \cite{vaswani2017attention} to vision-language (VL) tasks has revolutionized the field of VL modeling. In the space of only a few years, a huge array of multimodal (mostly vision) transformers have appeared in the literature. One thing that all of these models have in common is that their high parameter size translates to increasingly large compute and training data requirements to train them from end-to-end. At the smaller end are early encoder-only models such as VisualBERT, which contain around 100M parameters. This problem has become acute in recent years, as extremely large, commercial models require compute and electricity to function that make them commercially non-viable. 

Additionally, these "small" models are generally trained with very large datasets, often consisting of millions of unique images paired with text (Conceptual Captions \cite{sharma-etal-2018-conceptual} for instance) that are difficult to create, process and store. Furthermore, datasets are usually scraped from internet, often with text derived from ALT text captions, which may not translate to accurate depictions of the images. Datasets derived from human annotators, such as MSCOCO \cite{lin2014microsoft}, or with captions in low resource languages are naturally much smaller and more difficult to develop. 

In conjunction with practical constraints are considerations from child development. American children begin speaking, on average, at around 12 months of age \cite{gilkerson2017mapping} with some estimates that children only need to hear about 5 million words before they can utter recognizable words. While it is difficult to estimate how much visual stimuli children have been exposed to and how much of a child's visual experience is integrated into the linguistic system, it is not controversial to claim that the meaning of many words is 'grounded' into embodied experience, particularly visual experience \cite{harnard:grounding}. For example, color words or animal words refer to visually perceived entities. Visual language models are a step in the direction of bringing the visual and linguistic worlds together, but work needs to be done to make compact VL models that train and operate efficiently on developmentally plausible sized datasets. 

Creating compact and data-efficient transformer models trained on text has been an active area of research. Models such as DistilBERT \cite{sanh2019distilbert}, TinyBERT \cite{jiao2019tinybert} and MobileBERT \cite{sun2020mobilebert} are greatly reduced in size and compute requirements. And studies such as \citet{fields2023exploring} have shown that transformers can be both parameter and data efficient by simply training them from end-to-end. Unfortunately, there have to date been very few of these efforts with respect to VL transformers. The few studies that have so far been produced have primarily concentrated on using knowledge distillation, a process whereby the output of a large "teacher" model is used to train a smaller "student" model \cite{wang2022efficientvlm}. Inspired by distilled transformers such as DistilBERT \cite{sanh2019distilbert}, training models of this nature is both logistically and conceptually very complex. Other efforts have adopted the strategy of pruning, removing some proportion of the weights in a large pretrained model \cite{gan2022playing}. Both of these strategies rely however, on incorporating a full-size pretrained model at some point in the training process. For many researchers and practitioner of VL models, this is not desirable and confound a model's results when end-to-end training is indicated.  

In this study, we seek to address these gaps by investigating the training and evaluation of small-scale models for discriminative (i.e,. not generative) VL tasks in a low-resource setting. The goals of this study are two-fold. First, we hope to establish that small-scale transformers can perform well on VL tasks with small models, small training datasets, and limited compute. Secondly, we create and present a compact model resulting from these efforts that can be used for research or practical applications by researchers with limited resources. Our specific contributions can be summarized as follows:
\begin{itemize}
    \item We show that for compact VL transformers, with limited training data and compute, two-tower encoder models are more effective than one-tower encoders.\footnote{It is fairly common to refer to towers as \textit{streams} (e.g., a visual and a text stream), though in the literature both terms are used. We follow \citet{fields2023vision} and use the term \textit{tower} in this paper.}
    \item We show that we can create efficient VL models by incorporating convolutional vision models into a two-tower encoder architecture.
    \item We show that a variety of configurations to the cross-modal fusion layers of a two-tower encoder lead to the same downstream results.
    \item We present a compact VL model, using the convolutional image model EfficientNet-b2 \cite{tan2019efficientnet}, that can be trained end-to-end with relatively low compute and data inputs. 
\end{itemize}

The rest of this paper will proceed as follows: Section~\ref{sec:related-work} cover some of the relevant related work. Section~\ref{sec:data} will cover the datasets used for both for pretraining and evaluation. In Section~\ref{sec:experiment-1} we determine whether a one-tower or a two-tower encoder is best suited for parameter-efficient modeling. Section \ref{sec:experiment-2} describes our experiments for determining the best backbone architecture for our small-scale VL transformer. The third and final set of experiments, described in Section \ref{sec:experiment-3}, we investigate the other architectural elements of our compact VL transformer and discuss the results of our final model in Section \ref{sec:final-model}. Finally we close the paper with some brief concluding remarks in Section \ref{sec:conclusion}.

\section{Related Work}
\label{sec:related-work}

Though the field of VL modeling has expanded rapidly in recent years  (see \citet{fields2023vision} for an overview), there has to date been relatively little research on parameter-efficient VL modeling. As in the domain of natural language processing, most research has focused on compressing existing pretrained models using techniques such as knowledge distillation or pruning. DistilVLM \cite{fang2021compressing} and EfficientVLM \cite{wang2022efficientvlm} both use knowledge distillation much the way that models like DistilBERT \cite{sanh2019distilbert} do. Studies such as \citet{gan2022playing} on the other hand, attempt to compress VL transformers through pruning (i.e., removing parameters) while others use quantization techniques.

Works such as the MiniVLM model introduced in \citet{wang2020minivlm} attempt to create smaller models through training them end-to-end. This effort is closest to our work in spirit and practice. Though this model has many virtues it uses a separate object detector module to create visual features, substantially increasing the model's compute requirements and inference time. It was also trained with a very large dataset and doesn't entirely fit our definition of low-resource.

Finally, our research is directly related to the efforts of BabyLM challenge \cite{hu2024findings}. Similar to our work, \citet{Bunzeck2023-sl}, performed a systematic search of GPT-syle generative language models to find a compact model with few parameters.  More recently, \citet{Loaiciga2025-ie} explored lower batch sizes for training a small LM, and \citet{salhan-etal-2025-best} explored the best sequence length for small LMs. However, these models were only trained on text, whereas we are interested in language and vision. 

More recently, the BabyLM challenge has incorporated vision into training and evaluation on language tasks \cite{babylm-2025-main}; the most related work from those efforts most related to ours is \citet{Ganescu2025-sx} which introduced a VL architecture trained on small amounts of data, but was still appreciably larger and had a more complex training regime than we present here.

\section{Data and Evaluation Criteria}
\label{sec:data}

In this section, we explain the datasets used for all of the experiments we describe below. The section is broken up into three subsections, the first two dedicated respectively to pretraining data and data for fine-tuning and evaluation. For each dataset in each section we provide a brief description of the dataset as well as the rationale for including them in our experiments. In the final subsection we discuss the criteria by which we judge our efforts at model development to be successful.  

\paragraph{Pretraining Data}
\label{subsec:data-pretraining}

For pretraining, we have opted to make use of only two publicly available VL datasets: MSCOCO \cite{lin2014microsoft} and Visual Genome \cite{krishna2017visual}. Though there are other VL datasets available with significantly more image-text pairs (Conceptual Captions \cite{sharma-etal-2018-conceptual}, and Visual Question Answering \cite{agrawal2016vqavisualquestionanswering}, for example), we opted against using them to train our models. The reasoning here is twofold: Firstly, as the goal of our model is efficiency, we aim to train our model with a minimum of data where possible. The combined set of unique images in our two chosen datasets is fairly small, making the data easy to collect, store and process with limited resources. Secondly, we want to focus on reference resolution (MSCOCO) as our chosen visual task as it is a more common task that children engage in during their early language development \cite{McCune2008-oy}.


During pretraining, we combine MSCOCO and Visual Genome to produce a dataset containing just over 6 million image-text pairs, but only 180k unique images. After processing, the combined size of this dataset is less than 30GB, which makes it a relatively small VL pretraining dataset by almost any standard. 

\begin{figure}[]
\centering
\includegraphics[scale=0.25]{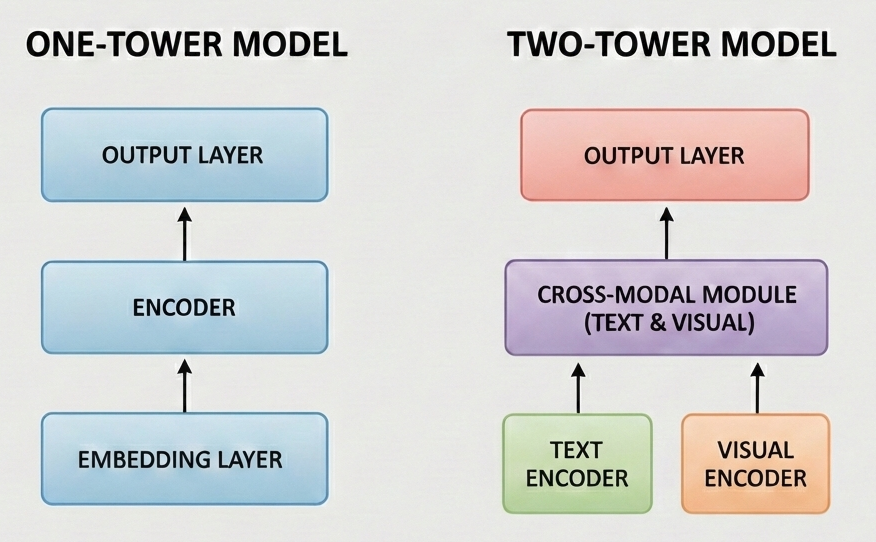}
\caption{Simple visual representation of two-tower and one-tower encoder models.}
\label{fig:one-vs-two}
\end{figure}

\paragraph{Evaluation Data}
\label{subsec:finetuning}

We have chosen a set of 3 tasks with which to fine-tune and evaluate the models from each of our experiments. The first of these, \textbf{NLVR2} \cite{suhr2018corpus} is a commonly used task that allows us to easily compare our results with previously published efforts. Further, it is a relatively challenging task that requires a model to answer a true or false question that relates two images. Though there are only two possible outcomes, the addition of second image also makes this quite a challenging task.

The second task, \textbf{SNLI-VE} \cite{xie2019visual} is somewhat less common and also less challenging. In this task, a model is presented with an image text-pair and must determine if the image entails a given sentence, contradicts the sentence or is neutral with respect to the sentence. Though less commonly used, the dataset is relatively small---fine-tuning and evaluation can be done in short order. This aids in the model development process by providing quick feedback. 

Our final downstream task is \textbf{reference resolution with the RefCOCO dataset} \cite{kazemzadeh2014referitgame}. Here, a model is presented with an image that is segmented into several objects and a sentence referring to one of these objects. The model must then determine which object the sentence is referring to (photos have an average of 15 objects to choose from). we include the reference task with RefCOCO for two reasons. Firstly, it is a fine-grained task that requires the model to specifically identify objects within an image. Secondly, the RefCOCO task is more directly aligned with child development as a visual task \cite{McCune2008-oy}. All three of these tasks use accuracy as their evaluation metric.

\paragraph{Model Desiderata} 
\label{subsec:criteria}
Compact models generally do not perform as well as their much larger counterparts, so we do not expect to obtain state-of-the-art results on any of our chosen benchmarks. Rather, we hope to obtain a compact model, in terms of number of parameters, that satisfies the requirements listed below:
\begin{itemize}
    \item Trainable end-to-end using a very small compute budget (in our case,  two NVIDIA l40s GPUs). The purpose here is to work towards models that are potentially more developmentally plausible, and require less electricty/carbon emissions to train and evaluate. 
    \item Performs on-par with other small VL models such as MiniVLM \cite{wang2020minivlm} on the NLVR2 task.
    \item Performs well on the reference resolution task with the RefCOCO dataset \cite{kazemzadeh2014referitgame}. Specifically, we aim to beat or match a score of 70\% obtained using the simple, yet effective Words as Classifiers (WAC) model \cite{schlangen2015resolving}. Other, recent state-of-the-art models exist for the RefCOCO benchmark, but they have appreciably more parameters and require more resources to train, so we opted for this baseline as a respectable baseline for our model to improve over. 
    \item Does not require additional pretraining data beyond MSCOCO and Visual Genome, our chosen datasets for our purposes.
\end{itemize}

\begin{table*}
\centering
\begin{tabular}{|l|l|l|l|l|l|l|l|l|}
 \hline
 \textbf{Model Type} &  \textbf{Text} & \textbf{Vision} & \textbf{Params} & \textbf{SNLI} & \textbf{NLVR2} & \textbf{Ref. Res.} & \textbf{Avg.} \\ 
 \hline
 \hline
 One-Tower & N/A & DeIT-S & 33.7M & 0.699 & 0.545 & 0.536 & 0.593\\
 \hline
 Two-Tower & ELECTRA-T & DeIT-T & 33.7M & \textbf{0.745} & \textbf{0.668} & \textbf{0.713} & \textbf{0.709}\\
 \hline
 \hline
  One-Tower & ELECTRA-B & N/A &  54.3M & 0.710 & 0.570 & 0.595 & 0.625 \\
 \hline
 Two-Tower & ELECTRA-S & Swin-S & 54.3M & \textbf{0.750} & \textbf{0.678} & \textbf{0.704} & \textbf{0.711}\\
 \hline
\end{tabular}
\caption{\label{table:one-vs-two}
\textbf{Results for One-Tower vs Two-Tower Study} All values are derived from the dev split for each task.
}
\end{table*}

\section{Experiment 1: One vs Two Tower Architecture}
\label{sec:experiment-1}

In this first experiment we ask, \textit{what type of architecture is best suited for small-scale modeling given our hardware constraints?} A fairly large range of architectures exist. However, either one-tower or two-tower encoders are generally best suited to our evaluation tasks, and they are usually more parameter efficient than decoder-based generative models. 
One-tower encoders consist of a single encoder that processes both image and text inputs together. Its architecture is very similar to NLP encoder models such as BERT \cite{devlin2019bert} with modifications to its embedding layer to accommodate both image and text features. Two-tower models, on the other hand, contain a separate vision and text encoder to handle the different inputs. The output of each encoder is then fed to a series of cross-attention layers that relate the two modalities and create a multi-modal output. A simple visual depiction of both model types can be found in Figure~\ref{fig:one-vs-two}.

Intuitively, the simplicity of the one-tower encoder architecture seems best suited to constructing a compact model as it has fewer modules. In order to test whether this intuition is correct, we conduct a systematic experiment to determine whether a one-tower or two-tower encoder is best suited to compact modeling using the procedure outlined below. The results of this experiment will guide the course of the two sets of experiments that follow and determine the basic architecture of the compact model that we seek to construct.

\paragraph{Experimental Setup and Procedure} This experiment, and those that follow, were conducted using the Renaissance VL modeling and evaluation harness \cite{fields2024renaissance}. Renaissance allows users to create novel encoder architectures by using text and vision models from the Huggingface Hub as modules. In order to make an accurate comparison between one-tower and two-tower architectures, we must construct models of each type which are as close to the same size as possible, in terms of number of parameters. Toward this end, we construct two pairs of models with each pair containing a one-tower encoder and a two-tower encoder of similar sizes. We use similar architectures within each model pair and match each encoder module's size as closely as possible, to ensure a fair comparison. The model pairs we use are summarized immediately below. 

In the first pair of models, each encoder module contains about 33.7 million parameters. 
\begin{itemize}
    \item \textbf{One-Tower}: The encoder module for this one-tower model is based on the DeiT-Small \cite{touvron2021training} and use patch embeddings \cite{dosovitskiy2020image} for image features  and BERT-style word-piece embeddings for text features \cite{devlin2019bert}. 
    \item \textbf{Two-Tower}: For the two-tower encoder, we use ELECTRA-Small \cite{clark2020electra} as the text encoder and DeiT-Tiny \cite{touvron2021training} as the vision encoder.  We adjust the dimensions of the cross-modal module to match the size of the one-tower model. In this case, our cross-modal module contains 7 layers with a hidden size of 256. 
\end{itemize}

In the second pair of models, each encoder module consists of around 54.3 million parameters.
\begin{itemize}
    \item \textbf{One-Tower}: The encoder module for the one-tower encoder is based on the ELECTRA-Base model \cite{clark2020electra}. We make use of the Renaissance platform to customize the model's dimensions to produce a model of the desired size. Our customized version of ELECTRA-Base contains 12 layers, each with a hidden size of 448. Naturally, this model contains randomly initialized weights rather than values from a pretrained model. However, researchers have previously found that this will not adversely affect downstream results \cite{fields2024renaissance}.   
    \item \textbf{Two-Tower}: The text encoder module is based on ELECTRA-Small \cite{clark2020electra} and the vision transformer Swin-Tiny \cite{liu2021swin}. We construct the cross-modal module to 7 layers with a hidden size of 256 to match the size of encoder in the one-tower model.
\end{itemize}

We pretrain both sets of models using standard masked language modeling and image-text matching using the dataset described in \ref{subsec:data-pretraining}.\footnote{The interested reader can find full descriptions of these pretraining tasks in \cite{fields2023vision}.} In our experience, pretraining generally produces the best results when training is conducted with the largest batch size that the training compute setup can accommodate. In our case, that is 704 for the first pair of models and 512 for the second pair of models. For both model pairs we use a learning rate of 1e-4 and train for 100k steps. 

\paragraph{Results} 

The results of this first set of experiments are summarized in Table~\ref{table:one-vs-two}. We see immediately that for both model pairs, the two-tower variation performs best on all of our downstream tasks. While the accuracy scores for the SNLI-VE task in both model pairs were separated by less than 5 percentage points, we see much large differences in the other two tasks. For both NLVR2 and Reference Resolution, we see double digit disparities between the models in each pair. While it is hard to make a direct comparison between models with different architectures, the size and range of the performance differences make this result fairly definitive. In our low-resource setting, using a small dataset and compact models, two-tower encoders outperform one-tower encoders by a fairly large margin on a variety of tasks. This matches a similar observation reported in \citet{fields2024renaissance}. For the remainder of our experiments, we consider the two-tower model as our base architecture.

\begin{table*}
\centering
\begin{tabular}{|l|l|l|l|l|l|l|}
 \hline
 \textbf{Text} & \textbf{Vision} & \textbf{Params} & \textbf{SNLI} & \textbf{NLVR2} & \textbf{Ref. Res.} & \textbf{Avg.} \\ 
 \hline
 \hline
 ELECTRA-T & DeIT-T & 22.6M & 0.728 & 0.618 & 0.712 & 0.686\\
 \hline
 ELECTRA-S & DeIT-T & 30.4M & 0.744 & 0.665 & 0.717 & 0.709 \\
 \hline 
 MobileBERT & DeIT-T & 41.5M & 0.742 & 0.665 & 0.645 & 0.684 \\
 \hline
 \hline
 ELECTRA-T & DeIT-S & 38.9M & 0.732 & 0.649 & 0.703 & 0.695 \\
 \hline
 ELECTRA-S & DeIT-S & 46.7M  & 0.747 & \textbf{0.684} & \textbf{0.723} & \textbf{0.718}\\
 \hline 
 MobileBERT & DeIT-S & 57.8M & \textbf{0.753} & 0.676 & 0.709 & 0.713 \\
 \hline
\end{tabular}
\caption{\label{table:backbone-1}
\textbf{Results for Part 1 of Backbone Architecture Study} using an image encoders with traditional transformer architectures. Parameter counts are for encoder modules without classification heads. All values are derived from the dev split for each task.
}
\end{table*}

\section{Experiment 2: Backbone Architecture Search}
\label{sec:experiment-2}

In our second set of experiments, we ask \textit{which pretrained language and vision models produce the best compact model, given our desiderata?} In the parlance of VL transformers, the pretrained vision and language modules are often referred to as the "backbone" architecture. Informed readers know that there are a daunting array of transformer models available for text processing and an increasingly large set of transformers for computer vision. However, the goal of a compact model can only be achieved by using suitably compact vision and language transformers.

Our results from Experiment 1 argue for the two-tower encoder architecture. Because a two-tower model has three separate encoder modules, this further narrows the choice of encoder models down to the smallest text and vision models available on the Huggingface model hub. We have assembled a list of candidate models that work with our evaluation harness and are sufficiently small in size. We can break these models into three distinct groups: candidate text transformers, candidate vision transformers and candidate vision models with other architectures. All of these models are listed below, with the number of parameters each contains:

\noindent
Candidate Text Modules:
\begin{itemize}
    \item ELECTRA-Tiny, 5.7M \cite{fields2023exploring}
    \item ELECTRA-Small, 13.7M \cite{clark2020electra}
    \item MobileBERT, 24.6M \cite{sun2020mobilebert}
\end{itemize}

\noindent
Candidate Transformer Image Modules:
\begin{itemize}
    \item DeIT-Tiny, 5.6M \cite{touvron2021training}
    \item DeIT-Small, 21.6M \cite{touvron2021training}
\end{itemize}

\noindent
Candidate Image Modules with Alternative Architectures:
\begin{itemize}
    \item Swin-Tiny, 27.5M \cite{liu2021swin}
    \item Resnet-50, 23.5M \cite{he2016deep}
    \item EfficientNet-b2, 7.7M \cite{tan2019efficientnet}
\end{itemize}

We elect to test only three text models as the compact text transformers are a well explored topic and there is generally less variation from model to model. We choose 2 candidate vision transformers: DeIT-Tiny and DeIT-Small are both small variations of the ViT model introduced in \citet{dosovitskiy2020image} and vary only in the size of the model. In the second set of vision models, we consider architectures other than the vanilla image transformer. Swin-Tiny is a compact hierarchal transformer that uses a shifted window technique to give the model better 2D reasoning \cite{liu2021swin}. Though the architecture is inspired by the transformer, it is specifically designed to deal with image processing. Resnet-50 and EfficientNet-b2 are both traditional convolutional neural network (CNN) models. To our knowledge, there is no two-tower model in the literature that uses a CNN as its image encoder. However, the Renaissance harness allows for their use and we examine their performance relative to image transformers. 


\begin{table*}
\centering
\begin{tabular}{|l|l|l|l|l|l|l|}
 \hline
 \textbf{Text} & \textbf{Vision} & \textbf{Params} & \textbf{SNLI} & \textbf{NLVR2} & \textbf{Ref. Res.} & \textbf{Avg.} \\ 
 \hline
 \hline
 ELECTRA-S & Swin-T & 52.5M & 0.719 & 0.637 & 0.698 & 0.684\\
 \hline
 ELECTRA-S & Resnet-50 & 48.8M & \textbf{0.752} & \textbf{0.687} & 0.631 & 0.690 \\
 \hline 
 ELECTRA-S & Efficientnet-b2 & 32.8M  & 0.749 & 0.677 & \textbf{0.719} & \textbf{0.715}\\
 \hline 
\end{tabular}
\caption{\label{table:backbone-2}
\textbf{Results for Part 2 of Backbone Architecture Study} using image encoders with alternative architectures. Parameter counts are for encoder modules without classification heads. All values are derived from the dev split for each task.
}
\end{table*}

\paragraph{Experimental Setup and Procedure}

We break this experiment into two parts. In Part 1, we train all possible combinations of traditional transformer image candidates and text candidates. The full cross-product of these candidate models results in 9 model variations to pretrain and evaluate. In order to isolate the contribution of the backbone architectures, we fix the dimensions of the cross-modal encoder layers. Each model contains 6 cross-modal layers with a hidden size of 256. Because we have such a large set of candidate models to train, we reduce the duration of training from 100k steps to 50k steps. In order to accommodate the largest of our candidate models, we also reduce the pretraining batch size to 512. We use a learning rate of 1e-4 and again train with standard masked language modeling and image-text matching tasks. Finally, we evaluate our models on the same three downstream tasks as we did in Experiment 1.

In Part 2, we train and evaluate our image encoders with architectures different than the traditional image transformer. In the interest of saving time and compute, we train only model variations using the most successful text model from the first set of experiments. This reduces the set of model variations from 9 to 3. We  use the same training settings as the first set. Of all 9 model variations from both sets of experiments, we choose the one with the best average score on our three downstream tasks. If multiple models are very close in average score, we use the one with the fewest parameters.

\paragraph{Results} 

The results for Part 1 are displayed in Table \ref{table:backbone-1}. We see that the model with ELECTRA-Small as its text encoder and DeIT-Small as its vision encoder performed the best overall. In terms of model size, this combination is the second-largest model variation that we tested with a total encoder size of 46.7M parameters. This is interesting as it indicates that though model size is definitely a factor, other factors play important roles. In almost every case, we see that the ELECTRA-Small model outperforms variations using the ELECTRA-Tiny and MobileBERT models, particularly on the challenging NLVR2 task. Of the vision models tested, DeIT-Small performed the best in each scenario. Given the superior performance of ELECTRA-Small, and with fewer parameters than MobileBERT, we opted to use it as the text model in the second set of experiments.

The results from Part 2, where we test alternative vision architectures are displayed in Table \ref{table:backbone-2}. Swin-Tiny, the hierarchical vision transformer, performed worse on average than both DeIT-Tiny and DeIT-Small from the previous set of experiments. This is somewhat surprising in that it is larger and often outperforms these models on pure computer vision tasks. However, the two convolutional models performed quite well. This is an exciting result, as to our knowledge, using CNNs for the vision encoder in a two-tower model is a novel and more parsimonious approach. ResNet-50 achieved the best score of any model on both the SNLI and NLVR2 tasks, but performed poorly on reference resolution which lowered its average score. The model using EfficientNet-b2 performed well on all tasks and led to an average score nearly identical to that of the DeIT-Small and ELECTRA-Small combination from the first set of experiments. Given that the model using EfficientNet-b2 is substantially smaller than the best model from Part 1, we use EfficientNet-b2 as the backbone vision model in our final set of experiments where we determine how to configure the multimodal fusion layer of our small model. 

\begin{table*}
\centering
\begin{tabular}{|l|l|l|l|l|l|l|l|}
 \hline
 \textbf{MHA} & \textbf{FFN} & \textbf{Layers} & \textbf{Params} & \textbf{SNLI} & \textbf{NLVR2} & \textbf{Ref. Res.} & \textbf{Avg.} \\ 
 \hline
 \hline
 256 & 1024 & 6 & 32.8M  & 0.749 & 0.677 & 0.719 & 0.715\\
 \hline
 \hline
 256 & 1024 & 4 & 29.1M & 0.749 & 0.672 & 0.720 & 0.714\\
 \hline
 256 & 1024 & 8 & 36.5M & 0.747 & 0.674 & 0.725 & 0.715\\
 \hline
 256 & 1024 & 10 & 40.2M & 0.746 & 0.673 & 0.721 & 0.713 \\
 \hline
 \hline
 192 & 768 & 6 & 27.8M & 0.747 & 0.663 & 0.712 & 0.705\\
 \hline
 320 & 1280 & 6 & 39.2M & \textbf{0.752} & \textbf{0.681} & 0.725 & \textbf{0.719}\\
 \hline
 384 & 1536 & 6 & 60.7M & 0.733 & 0.667 & \textbf{0.727} & 0.709\\
 \hline
\end{tabular}
\caption{\label{table:auxilary-1}
\textbf{Results Auxiliary Architecture Study} Parameter counts are for encoder modules without classification heads. All values are derived from the dev split for each task.
}
\end{table*}

\section{Experiment 3: Auxiliary Architecture Search}
\label{sec:experiment-3}

In the final set of experiments we ask \textit{whether or not we can improve our model's performance by altering the dimensions of its cross-modal encoder module}. As depicted in \ref{fig:one-vs-two}, the fusion encoder of a two-tower model consists of a dual stack of transformer layers that use cross-attention. Each transformer layer consists of a multi-head attention (MHA) mechanism followed by a simple feed-forward network (FFN). The MHA component of each layer in the fusion encoder uses cross-attention to combine the visual and textual streams. The FFN transforms the output of the MHA to an intermediate dimension, and then transforms the output back to the MHA's hidden size before its output is sent to the next layer. The size of each dimension can be altered and we observe what effect that has on our model's downstream performance. 


\paragraph{Experimental Setup and Procedure} The approach in this experiment is quite similar to that of the above experiments. However, here we fix the backbone architecture to that determined above and test the effect of varying the dimensions of the each model variation's fusion encoder. Per our results from Experiment 2, we use EfficientNet-b2 as our image encoder and ELECTRA-Small as our text-encoder. The fusion encoders for all models from Experiment 2 had an MHA size of 256, an FFN size of 1024 and 6 layers. The results obtained using these settings from experiment two serve as a baseline against which we compare the models from this experiment.

In most transformer models, the intermediate layer of the FFN is always four times larger than the size of the MHA and we follow that convention in this experiment. This means that we examine two model dimensions: width (the number of parameters in the MHA and FFN) and depth (the number of layers in the encoder). The space of possible configurations in this case is very large and we can only explore a small part of it. However, because the overall model must still be small enough to train with our limited compute setup, the fusion encoder must still be fairly small in size. In total, we train and evaluate 6 models. We begin exploring model width, then increase the number of layers from 6 to 8, 10 and 12 and also train one model with 4 layers to explore the effect that decreasing the depth of the encoder has. We then examine ranging the width of the encoder from an MHA size 256 to 192, 320 and 384. 

We again pretrain each model variation for 50k steps with a learning rate of 1e-4, using masked language modeling and image-text matching. We evaluate these models using the same 3 downstream tasks as we did in the previous experiments. As in Experiment 2, we choose the model with the highest average score. If two models have similar scores we choose the configuration with the fewest parameters. To complete the study, we train this final model configuration for an extended number of steps (150k) and evaluate the results.

\begin{table*}
\centering
\begin{tabular}{|c|c|c|c|c|c|c|}
 \hline
 \textbf{Model} & \textbf{Params} & \textbf{Training} & \textbf{SNLI} & \textbf{NLVR2} & \textbf{Ref. Res.} & \textbf{Avg.} \\ 
 \hline
 \hline
 Efficient & 39.2M  & 20 Epochs & 0.751 & 0.711 & .727 & 0.730\\
 \hline 
 MiniVLM & 52M & 100 Epochs & - & 0.737 & - & -\\
 \hline
 PixelBERT & 110M & 40 Epochs & - & 0.717 & - & -\\
 \hline
 WAC & - & - & - &  - & .700 & -\\
 \hline
\end{tabular}
\caption{\label{table:backbone-3}
\textbf{Results for Part 2 of Backbone Architecture Study} using image encoders with alternative architectures. Parameter counts are for encoder modules without classification heads. All values are derived from the dev split for each task.
}
\end{table*}

\paragraph{Results} The results for Experiment 3 are summarized in Table \ref{table:auxilary-1}. They indicate that the size of the fusion encoder doesn't influence the model's downstream performance as much as might be expected, at least for models of this size. Recall that the baseline configuration we used was 6 layers each with an MHA hidden size of 256, the results for this layer are displayed in the first row of the table. Altering the number of layers, either increasing or decreasing the number, had very little effect on the model's downstream performance. Changing the model's MHA size did produce more effect. Reducing the MHA size to 192 lowered the model's average performance an entire percentage point. Increasing the MHA size to 320 resulted in a small increase in average accuracy. Though the overall increase is very small, .004 increase in average score, all three tasks did improve over the baseline. Increasing the MHA size to 384 actually lowered average accuracy.

We saw in Experiment 2 that the size and architecture of vision and text encoders have a significant impact on the model's downstream performance. It seems unlikely that the size and configuration of the model's fusion encoder simply doesn't matter. Our results do indicate, however, that for a given model a significant range of possible configurations for fusion encoders lead to the same result. In future studies, testing this effect over a range of models with different backbone architectures, additional pretraining data and batch sizes would shed more light on this phenomenon. That said, the results here suggest that increasing the MHA size of the fusion encoder may lead to a modest increase in performance and the resulting model is still quite small, less than 40M parameters. As such, we train a final model with this configuration for an extended number of steps (150k) and discuss in the next section.

\section{ESsEN: Our Final Model}
\label{sec:final-model}

The preceding experiments have led us to a two-tower encoder model that uses EfficientNet-b2 as its image encoder and ELECTRA-Small as its text encoder. Its cross-modal fusion encoder has 6 layers, each with a hidden size of 320. Though our training dataset consisted of only 6M, image-text pairs and only 180k unique images, we were still able to achieve most of our desiderata set out in Section \ref{subsec:criteria}. We call this model ESsEN (\textbf{E}lectra-\textbf{S}mall \textbf{s}upported by \textbf{E}ffecient\textbf{N}et) and it has the virtue of being much smaller (39.2 million parameters total, from both the text and visual towers combined) and more efficient than other compact VL transformers currently available, see Table~\ref{table:backbone-3}. Although our performance on NLVR2 is 2 points short of our goal on the NLVR2 task, we compare it to MiniVLM \cite{wang2020minivlm} trained on much larger dataset. When we compared to PixelBERT, the only model in the literature we know of that is trained on the same dataset, we obtain similar results despite our model being a fraction of the size. We hope that compact VL models of this nature will make VL research more accessible as the field continues to mature.\footnote{We will make the model weights available upon publication.}

\section{Conclusion}
\label{sec:conclusion}

In service of making VL research more accessible, we have shown that we can obtain useful results from small VL models, with limited data and limited compute, and using data that follows more closely from child development in that it uses vision-language data in a reference resolution task (i.e., RefCOCO). In the first series of experiments, we showed that in this minimalist setting, two-tower encoders perform better than one-tower encoders. In the second set of experiments we showed that compact convolutional models like EfficientNet can improve parameter efficiency over similarly sized transformer models. In the third and final experiment, we show that a variety of configurations of a two-tower encoder's fusion module will lead to the same downstream results. Finally, we present ESsEN, a compact two-tower encoder that can be trained with limited compute and limited training data. The field of VL modeling is rapidly evolving and we hope that research tools such as ESsEN and the Renaissance modeling platform \cite{fields2024renaissance} will open the field to a wider variety of researchers.

For future work, we will continue to explore compact VL models that can be trained on smaller amounts of data, yet produce models that are useful, and models that can be trained not only on pre-existing datasets, but can learn in real time interactive settings where language and vision are both modalities used for model learning.

\section*{Limitations}

Though our choice to use limited resources made sense for this study, it can also be viewed as a limitation. Larger pretraining datasets and and more compute would almost certainly have led to better scores on the various benchmarks that we employed, even if the model's size was held steady. Future studies to maximize the performance of small models like the ones presented here would likely make a significant contribution to the field.

Moreover, this paper focused only on discriminative tasks, whereas research has shifted focus to generative tasks. Generative tasks definitely are important, and a lot of effort is being put into their research and development, but there are many valid and important uses of discriminative VL models for research and industrial purposes that would cut carbon emissions, if adopted. 

\section*{Ethics Statement}

We have no ethical considerations or conflicts of interest to report.


\bibliography{lrec2026-example}
\label{sec:reference}


\end{document}